\documentclass{article}

\usepackage[final]{AdvML_Frontiers_2024}

\usepackage[utf8]{inputenc} 
\usepackage[T1]{fontenc}    
\usepackage{url}            
\usepackage{booktabs}       
\usepackage{amsfonts}       
\usepackage{nicefrac}       
\usepackage{microtype}      
\usepackage{xcolor}         

\usepackage{graphicx}
\usepackage{algorithm}
 \usepackage{algpseudocode}  
\usepackage{multirow}
\usepackage{booktabs}
\usepackage{amsmath} 

\definecolor{cvprblue}{rgb}{0.21,0.49,0.74}
\definecolor{cvprred}{rgb}{0.74,0.21,0.49}
\usepackage[breaklinks,colorlinks,citecolor=cvprblue,linkcolor=cvprred]{hyperref}

\title{TrackPGD: Efficient Adversarial Attack using Object Binary Masks against Robust Transformer Trackers}

\author{%
  Fatemeh Nourilenjan Nokabadi\\ 
  Université Laval, IID, Mila\\
  \texttt{fatemeh.nourilenjan-nokabadi.1@ulaval.ca}\\
  \And
  Yann Batiste Pequignot\\
  Université Laval, IID\\
  \texttt{yann.pequignot@iid.ulaval.ca} \\
  \AND
  Jean-Fran\c{c}ois Lalonde\\
  Université Laval, IID\\
  \texttt{jean-francois.lalonde@gel.ulaval.ca} \\
  \And
  Christian Gagné\\
  U. Laval, IID, Mila, Canada-CIFAR AI Chair\\
  \texttt{christian.gagne@gel.ulaval.ca} \\
}

\begin{document}

\maketitle

\begin{abstract}
  Adversarial perturbations can deceive neural networks by adding small, imperceptible noise to the input. Recent object trackers with transformer backbones have shown strong performance on tracking datasets, but their adversarial robustness has not been thoroughly evaluated. While transformer trackers are resilient to black-box attacks, existing white-box adversarial attacks are not universally applicable against these new transformer trackers due to differences in backbone architecture. In this work, we introduce TrackPGD, a novel white-box attack that utilizes predicted object binary masks to target robust transformer trackers. Built upon the powerful segmentation attack SegPGD, our proposed TrackPGD effectively influences the decisions of transformer-based trackers. Our method addresses two primary challenges in adapting a segmentation attack for trackers: limited class numbers and extreme pixel class imbalance. TrackPGD uses the same number of iterations as other attack methods for tracker networks and produces competitive adversarial examples that mislead transformer and non-transformer trackers such as MixFormerM, OSTrackSTS, TransT-SEG, and RTS on datasets including VOT2022STS, DAVIS2016, UAV123, and GOT-10k.
\end{abstract}

\section{Introduction}
\label{sec:intro}

\begin{figure}
     \centering
     \centerline{\includegraphics[width=\textwidth]{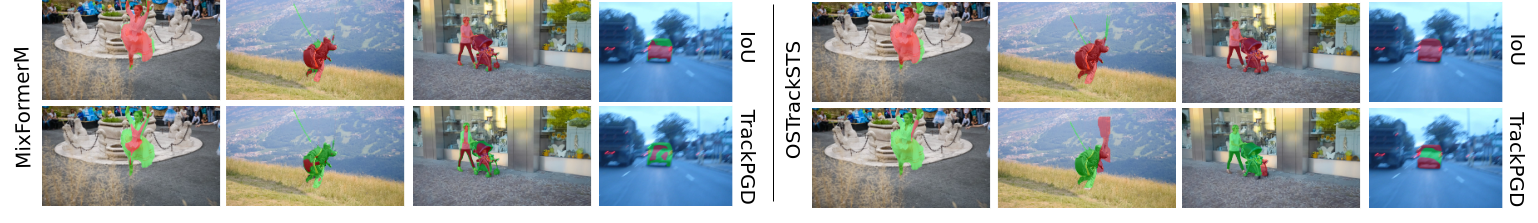}}
     \caption{The binary masks generated by MixFormerM~\citep{cui_MixFormer_2022} and OSTrackSTS~\citep{ye_joint_2022} attacked by our TrackPGD vs. IoU~\citep{jia_iou_2021} which is the most recent and applicable attack for these two trackers up to now. The green/red masks are the before/after the attack outputs.}
     \label{fig:intro}
\end{figure}

Adversarial attacks~\citep{guo_spark_2020,jia_iou_2021,jia_robust_2020,yan_cooling-shrinking_2020} can significantly harm neural networks performances by adding carefully crafted imperceptible noise to the input. For instance, classification networks are highly susceptible to such adversarial perturbations~\citep{shao_adversarial_2022,mahmood_robustness_2021}. For object segmentation and object tracking, some recent attacks~\citep{gu_segpgd_2022,guo_spark_2020,jia_iou_2021} have been developed to disrupt the object masks and bounding boxes output, respectively. Since 2020, the Visual Object Tracking (VOT) challenge~\citep{kristan_eighth_2020} introduced a new sub-challenge that requires object trackers to present binary masks as output and be evaluated based on their mask generation ability. In addition, novel trackers with transformer backbones have emerged~\citep{cui_MixFormer_2022, ye_joint_2022, chen_high-performance_2023,bhat_learning_2020}, whose approaches infer the object mask, bounding box, and confidence score as a triple output. Given that objects with complex shapes are rarely aligned with the frame axis, the corresponding axis-aligned bounding boxes contain more background pixels than object pixels~\citep{paul_robust_2022}. Therefore, object bounding boxes are not the best option to describe an object's appearance in comparison to object masks. As a result, object binary masks have become increasingly important both as a valid output of trackers and a valid criterion for evaluation in tracking benchmarks. Visual tracking can be compromised by adversarial attacks in different ways, by manipulating elements such as object bounding boxes~\citep{jia_iou_2021}, classification labels~\citep{jia_robust_2020,yan_cooling-shrinking_2020,guo_spark_2020}, and regression labels~\citep{jia_robust_2020,guo_spark_2020}.  In order to apply white-box attacks on new transformer trackers~\citep{cui_MixFormer_2022, ye_joint_2022, chen_high-performance_2023,bhat_learning_2020}, the adversarial loss should be computed using the tracker's elements and outputs. While some white-box attacks, such as SPARK~\citep{guo_spark_2020} and RTAA~\citep{jia_robust_2020}, are not applicable to new robust object trackers, such as RTS~\citep{paul_robust_2022} and MixFormer~\citep{cui_MixFormer_2022}, the black-box attacks are not effective enough to challenge these trackers. In this work, we propose TrackPGD, a novel and efficient white-box attack that uses the object binary mask as a proxy. Figure~\ref{fig:intro} illustrates TrackPGD's effectiveness in disrupting binary masks of MixFormerM~\citep{cui_MixFormer_2022} and OSTrackSTS~\citep{ye_joint_2022}, with green/red masks representing predictions before/after the attack. TrackPGD caused more damage than the Intersection over Union (IoU) method~\citep{jia_iou_2021}, which only slightly degraded the bounding box while leaving the binary mask mostly unchanged. The IoU method is the most recent applicable attack against MixFormerM~\citep{cui_MixFormer_2022}, OSTrackSTS~\citep{ye_joint_2022} and RTS~\citep{paul_robust_2022}.

\begin{table*}
    \centering 
    \caption{Characterization of attack methods-- IoU~\citep{jia_iou_2021}, CSA~\citep{yan_cooling-shrinking_2020}, SPARK~\citep{guo_spark_2020}, RTAA~\citep{jia_robust_2020} and TrackPGD (ours)-- on robust object trackers including MixFormerM~\citep{cui_MixFormer_2022}, OSTrackSTS~\citep{ye_joint_2022}, TransT-SEG~\citep{chen_high-performance_2023} and RTS~\citep{paul_robust_2022}. TrackPGD is the only applicable White-box attack against MixFormerM, OSTrackSTS and RTS, using a new proxy (i.e. binary mask), while the two other methods that may be applied to these trackers, IoU and CSA, are Black-box methods manipulating bounding box.}
    \label{tab:proxy}
    \resizebox{\linewidth}{!}{
    \begin{tabular}{lllcccc}  \toprule 
    Attack Setting & Method & Attack Proxy  & MixFormerM & OSTrackSTS & TransT-SEG & RTS \\
    \midrule 
    \multirow{2}{*}{Black-box}  
      & IoU & Object bbox & \checkmark & \checkmark & \checkmark & \checkmark \\
      & CSA &  Object bbox, heat-maps & \checkmark & \checkmark & \checkmark & $\times$\\ \midrule
      \multirow{3}{*}{White-box}  
     & SPARK & Regression and classification labels &  $\times$ & $\times$  & \checkmark & $\times$ \\
     & RTAA & Regression and classification labels & $\times$  & $\times$  & \checkmark & $\times$\\
     & \textbf{TrackPGD} &\textbf{Object binary mask} & \checkmark & \checkmark & \textbf{\checkmark} & \checkmark \\
    \bottomrule \\
    \end{tabular}}
\end{table*}


Targeting object binary masks in adversarial attacks is essential due to their increasing use in tracking backbones~\citep{cui_MixFormer_2022,ye_joint_2022,yan_alpha-refine_2021,paul_robust_2022} and datasets~\citep{kristan_tenth_2023}. In Table~\ref{tab:proxy}, several attack methods and their attack proxy are listed along with their applicability to some new transformer trackers, MixFormerM~\citep{cui_MixFormer_2022}, OSTrackSTS~\citep{ye_joint_2022} and TransT-SEG~\citep{chen_high-performance_2023}. Using a novel attack proxy, the object binary mask, TrackPGD is developed to generate the adversarial frame patches from the predicted binary mask. In addition, the experimental results on object bounding box evaluation demonstrate that the TrackPGD is an effective attack over neural network-based object trackers. Table~\ref{tab:proxy} lists various attacks, the proxies, and applicability to new trackers. For example, MixFormerM~\citep{cui_MixFormer_2022} does not output object candidate labels, making label-based attacks such as RTAA~\citep{jia_robust_2020} and SPARK~\citep{guo_spark_2020} unsuitable as white-box attacks. In a nutshell, our contributions can be summarized as follows:
\begin{itemize}
    \item We propose a novel white-box attack, TrackPGD, which builds the adversarial noise from the binary mask to attack transformer trackers.
    \item A new loss term -- built upon SegPGD attack -- is proposed to compute the TrackPGD loss with the aim of misleading visual trackers in providing an accurate binary mask.  
    \item We demonstrate that MixFormerM suffers significant accuracy ($-75\%$), average overlap ($-97\%$) and robustness ($-91\%$) reductions after applying TrackPGD, evaluated on the VOT2022STS dataset. 
    \item Experimental results also demonstrate that the perturbations generated by TrackPGD have a significant influence on bounding box predictions in tracking benchmarks.  
\end{itemize}

\section{Related works}

\noindent\textbf{Robust Object Trackers}\quad Tracker robustness was initially defined as the frequency of failures, marked by the overlap between prediction and ground truth dropping to zero. This definition is employed in datasets like VOT2016~\citep{kristan_visual_2016} and VOT2018~\citep{kristan_sixth_2018}. A newer protocol introduced in VOT2020~\citep{kristan_eighth_2020} includes multiple anchors in each test sequence spaced approximately 50 frames apart to facilitate anchor-based short-term tracking~\citep{kristan_eighth_2020, kristan_tenth_2023}. In contrast, other tracking datasets, like LaSOT benchmark~\citep{fan_lasot_2019}, conduct a one-pass evaluation. The MixFormerM~\citep{cui_MixFormer_2022} was one of the three most robust trackers on the VOT2022STS~\citep{kristan_tenth_2023}. The MixFormer~\citep{cui_MixFormer_2022} tracker proposed a Mixed Attention Module (MAM) to jointly extract features of search and template regions. MixFormerM~\citep{cui_MixFormer_2022, kristan_tenth_2023} contains two networks, one tracker network and the other is AlphaRefine~\citep{yan_alpha-refine_2021} for object segmentation. But TransT-SEG~\citep{chen_high-performance_2023} jointly predicts the object bounding box and generates the object binary mask via a single network that uses a light cross-attention mechanism to fuse deep features. Another transformer tracker, OSTrackSTS~\citep{ye_joint_2022}, gained the third rank on the VOT-STS2022~\citep{kristan_tenth_2023} sequestered dataset. The OSTrackSTS~\citep{ye_joint_2022} is the One-Stream Track method~\citep{ye_joint_2022} coupled with the AlphaRefine~\citep{yan_alpha-refine_2021} to provide object binary masks. The Robust Tracking by Segmentation~\citep{paul_robust_2022} is a new robust discrimintative tracker to infer the object binary mask from a whole video frame. Using a segmentation branch, and an instance localization branch, RTS achieved a robust tracking performance by predicting the target mask in the tracking process~\citep{kristan_tenth_2023}.

\noindent\textbf{Adversarial attacks for object trackers}\quad The attack proxy can be classification labels~\citep{jia_robust_2020,yan_cooling-shrinking_2020,guo_spark_2020}, regression output~\citep{jia_robust_2020} or even the motion features of the object~\citep{jia_iou_2021,jia_robust_2020}. In the Robust Tracking against Adversarial Attacks (RTAA) method~\citep{jia_robust_2020}, specific labels for classification and regression are modified. An online incremental attack called SPARK~\citep{guo_spark_2020} has been developed for object trackers that use the previous frame perturbations and manipulate the object classification and regression labels to generate the adversarial frame. The Cooling-Shrinking Attack (CSA) method, as described in \citep{yan_cooling-shrinking_2020}, involves training two Generative Adversarial Networks (GANs) to create adversarial images for the SiamRPN++ tracker~\citep{li_siamrpn_2019}. The cooling term cools down the heated pixels in the heatmap of SiamRPN++, while the shrinking term shrinks the predicted bounding box. The IoU attack is a black-box approach that uses the predicted bounding box to inject two types of noises~\citep{jia_iou_2021} into frames. The directional and orthogonal perturbations are injected in a given frame to degrade the predicted bounding box from the previous prediction. 


\section{Preliminaries: SegPGD}


The SegPGD~\citep{gu_segpgd_2022} loss for a segmentation network ($L_\text{SegPGD}$) is defined to steer the neural network toward providing an inaccurate segmentation map. This loss is computed between the mask annotation $Y$ and the adversarial segmentation map $f_\text{seg}(X^\text{adv})$ generated by the segmentation network from the adversarial image $X^\text{adv}$:
\begin{equation}
L_\text{SegPGD}(f_\text{seg} (X^\text{adv}) , Y)  = 
\frac{1 - \lambda}{H W} L_\text{pos} +  \frac{\lambda}{H W} L_\text{neg} \,,
\end{equation}
where $\lambda$ is a coefficient that dynamically changes over iterations and $H$ and $W$ are the frame height and width, respectively. Here, $L_\text{pos}$ and $L_\text{neg}$ denote the cross-entropy loss restricted to correctly classified and wrongly classified pixels, respectively. The motivation for this reweighting stems from the observation that gradients of cross-entropy for wrongly classified examples tend to have larger amplitude, while the attack aims primarily at correctly classified pixels~\citep{gu_segpgd_2022}. 

To adapt SegPGD for use against object trackers and to target the predicted binary masks, we must overcome two key challenges. Firstly, while SegPGD operates on multi-class segmentation networks, binary masks only have two classes, necessitating the switching of classification labels between pixels. Secondly, the labels in the majority of samples are imbalanced, with the number of object pixels being significantly fewer than background pixels. To address these hurdles, we introduce TrackPGD as a white-box attack for object trackers which predict binary masks.

\section{Proposed method: TrackPGD} 

Let us formalize object tracking as using a tracker $\mathcal{F}(\cdot)$ that processes a video sequence $\mathcal{V} = \{ I_0, \ldots, I_{L}\}$ with $L+1$ frames, having the first frame object mask $M_0$ provided to initialize the tracker. The transformer tracker is expected to estimate the object mask $M_{\tau}$ at each time step $\tau\in\{1,\ldots,L\}$ from the given video frame $I_{\tau}$. The goal of the attack method is thus to create an adversarial video frame $I^{\text{adv}}_{\tau}$ at each time step to mislead the object tracker in its mask prediction.  The transformer tracker predicts at each time $\tau$ a mask $M_\tau$ which assigns to each pixel a probability of being an object. The TrackPGD attack aims to mislead the tracker in making these assignments by creating the adversarial frame $I^{\text{adv}}_{\tau}$. For object tracking, the mask annotation $G_\tau$  consists of two labels (object (1) and background (0)). It follows that finding adversarial examples can be achieved at optimization step $t$ both by maximizing $L_\text{SegPGD}(M^t, G_{\tau})$ or by minimizing $L_\text{SegPGD}(M^t, 1-G_{\tau})$, where we consider the mask $M_{\tau-1}$ predicted on the previous frame as the ground truth $G_\tau$ at step $\tau$. In the TrackPGD framework, we therefore decide to combine these two complementary objectives into a difference loss $L_{\Delta}$ as follows:
\begin{equation}\label{eq2}
L_{\Delta} = L_\text{SegPGD}(M^t, G_{\tau}) - L_\text{SegPGD}(M^t, 1-G_{\tau}) \,.
\end{equation}

Since the object pixels and background pixels are imbalanced in the video frame, we employed the focal loss~\citep{lin_focal_2020}. It is a useful loss for training highly imbalanced data, which is the case, here, for the object and background pixels. For TrackPGD, instead of using cross-entropy as in the original equation of focal loss~\citep{lin_focal_2020}, we use the $L_{\Delta}$ loss as follows:
\begin{align} \label{Eq3}
p_t & = G_{\tau} \mathcal{P} + (1 - G_{\tau})  (1 - \mathcal{P})\nonumber\\
L_\text{focal} & = \alpha_t (1 - p_t)^{\gamma} L_{\Delta} \,,
\end{align}
where $\mathcal{P}$ is a probability map derived from the application of the sigmoid function on the predicted binary mask $M_t$ (before thresholding) and $\gamma$ is the focusing parameter. This $\alpha$-balanced variant of the focal loss used is similar to the TransT-SEG~\citep{chen_high-performance_2023} except that $L_{\Delta}$ is used instead of cross-entropy. Then, we compute the dice loss~\citep{milletari_v-net_2016} between $M^t$ and $G_\tau$ as follows: 
\begin{equation}\label{Eq4}
     L_\text{dice} = 1 - 2\text{IoU}(M^t, G_{\tau}),
 \end{equation}
where $\text{IoU}$ is the Intersection over Union loss between the ground truth $G_{\tau}$ and the predicted binary mask $M^t$. The combination of focal loss (Equation~\ref{Eq3}), and dice loss (Equation~\ref{Eq4}), is used to compute the TrackPGD loss as follows:
\begin{equation}
\label{eq4}
L = \lambda_1 L_\text{focal} + \lambda_2 L_\text{dice} \,,
\end{equation}
where the $\lambda_1$ and $\lambda_2$ balance the focal and dice terms. 

For a detailed explanation of the TrackPGD algorithm and additional studies, please refer to the Supplementary materials.

\section{Experiments}
\label{sec:experiments}




\paragraph{Datasets.} Several video segmentation and tracking datasets. including DAVIS 2016~\citep{perazzi_benchmark_2016}, VOT2022STS~\citep{kristan_tenth_2023}, VOT2018~\citep{kristan_sixth_2018}, GOT10k~\citep{huang_got-10k_2021}, and UAV123~\citep{mueller_benchmark_2016}, are adopted in our experiments.

\paragraph{Victim Trackers.} We selected three popular transformer-based trackers, MixFormerM~\citep{cui_MixFormer_2022}, OSTrackSTS~\citep{ye_joint_2022}, TransT-SEG~\citep{chen_high-performance_2023} and one non-transformer tracker, RTS~\citep{paul_robust_2022}, to assess their adversarial robustness against adversarial attacks. 

\paragraph{Adversarial Attacks.} We applied four attack methods including CSA~\citep{yan_cooling-shrinking_2020}, SPARK~\citep{guo_spark_2020}, RTAA~\citep{jia_robust_2020}, and IoU~\citep{jia_iou_2021}, against these trackers. The TrackPGD perturbs the search region of the segmentation network in MixFormerM~\citep{cui_MixFormer_2022} and OSTrackSTS~\citep{ye_joint_2022} while the search region of the tracking network remains untouched. The hyperparameters $(\lambda_1, \lambda_2)$ of the TrackPGD loss are determined through the validation experiment on the DAVIS2016~\citep{perazzi_benchmark_2016}, as explained in the section Fine tuning the parameters.

\paragraph{Metrics.} The expected average overlap (denoted as EAO and AO), robustness, accuracy and AUC from VOT2022 metrics, Mean Jaccared Index $J(M)\%$ from DAVIS2016 evaluation toolkit, AUC and precision rates for GOT01k and UAV123 datasets are used to rank the attack performance.

\subsection{Binary mask assessment} 

\paragraph{Quantitative evaluation.} We evaluated the TrackPGD attack compared to other methods on two transformer trackers -- MixFormerM~\citep{cui_MixFormer_2022} and OSTrackSTS~\citep{ye_joint_2022} -- using the VOT-STS2022 dataset, see Table~\ref{tab:votMT}. Two streams of experiments from VOT-STS2022 (baseline and unsupervised) are conducted to indicate our attack performance under various protocols and metrics. Our findings demonstrate that TrackPGD was the most effective attack, reducing the robustness of MixFormerM~\citep{cui_MixFormer_2022} by around $91\%$ in the baseline experiment of VOT-STS2022 evaluation. TrackPGD also caused higher degradations to other evaluation metrics -- EAO, accuracy, and AUC -- than the two attack methods (i.e., IoU~\citep{jia_iou_2021} and CSA~\citep{yan_cooling-shrinking_2020} methods) against MixFormerM~\citep{cui_MixFormer_2022}, both on baseline and unsupervised streams of experiments. Among all measured metrics, TrackPGD ranked first by a significant margin on the VOT-STS2022 dataset. In attacking the OSTrackSTS tracker, TrackPGD also ranked first as an applicable attack per EAO, accuracy and AUC metrics. However, the IoU method~\citep{jia_iou_2021} performs better than our method on the robustness metric. To check further examples of mask prediction after attacks, we refer the reader to the Supplementary materials.

\begin{table*}
    \centering
    \caption{The MixFormerM~\citep{cui_MixFormer_2022}, OSTrackSTS~\citep{ye_joint_2022} results in generating object binary masks, when attacked by different methods on the VOT-STS2022~\citep{kristan_tenth_2023} dataset. TrackPGD is the only applicable white-box attack on MixFormerM and OSTrackSTS trackers. Bold/italic are the first/second ranks, respectively. The other white-box attacks, RTAA and SPARK are not applicable to these trackers due to difference in backbones. For further information, please check the text.}
    \label{tab:votMT}
    \resizebox{\linewidth}{!}{
    \begin{tabular}{p{0.15\textwidth}lcccccccc} 
    \toprule
     & & \multicolumn{3}{c}{Baseline} & Unsupervised & \multicolumn{4}{c}{Drop $\%$} \\
    \midrule
      Tracker & Attack  & EAO & Accuracy  & Robust. & AUC & EAO & Accuracy  & Robust. & AUC \\
    \midrule 
    \multirow{4}{*}{MixFormerM}
         & No Attack & 0.59 & 0.80 & 0.88  & 0.72 &  -- & -- & -- & -- \\
         & CSA & 0.56 &  0.80 &  0.86 & 0.65 &  4.58 & -0.63 & 2.61 & 9.40 \\
         & IoU & \textit{0.36} & \textit{0.66} & \textit{0.68} & \textit{0.51} &  
         \textit{39.05} & \textit{17.29} & \textit{23.07} & \textit{15.40} \\
        & \textbf{TrackPGD} & \textbf{0.02} &\textbf{ 0.20} & \textbf{0.08} &  \textbf{0.06} & \textbf{96.94} & \textbf{75.44} & \textbf{91.14} & \textbf{91.70} \\ 
        \midrule 
        \multirow{4}{*}{OSTrackSTS}  
         & No Attack & 0.59 & 0.77 & 0.87 &  0.66 &  - & - & - & - \\ 
         & CSA & 0.53 & 0.76 & \textit{0.82} &  0.62 &  10.54 & 2.32 & \textit{5.29} & 5.79 \\
         & IoU & \textit{0.31} & \textit{0.71} & \textbf{0.60} & \textit{0.48} &  \textit{47.45} & \textit{8.13} & \textbf{30.92} & \textit{27.94} \\
        & \textbf{TrackPGD} & \textbf{0.23} & \textbf{0.31} & 0.86 & \textbf{0.27}  &  \textbf{60.34} & \textbf{59.74} & 0.80 & \textbf{59.06} \\ 
       \bottomrule
    \end{tabular}}
\end{table*}

\subsection{Bounding box assessment}

\paragraph{Quantitative evaluation.} Table~\ref{fig:box} presents the attack results on the TransT-SEG~\citep{chen_high-performance_2023} and RTS~\citep{paul_robust_2022} trackers using the GOT-10k dataset~\citep{huang_got-10k_2021}. Although our method focuses on manipulating object binary masks, the reported numbers for bounding box evaluation are very close to those of competitive attacks aimed at misleading trackers into providing inaccurate bounding boxes. We surpassed the RTAA attack on the GOT-10k dataset in both success rate scores. However, RTAA performs better in terms of average overlap (AO). TrackPGD scores are also very close to SPARK scores, the most competitive attack using object bounding boxes on GOT-10k~\citep{huang_got-10k_2021}, in misleading trackers to predict accurate object bounding boxes. Despite our method's aim to misclassify object and background pixels, it proves to be a highly competitive approach compared to bounding box-centered attacks. Furthermore, our TrackPGD has an extra advantage over other white-box attacks which is that other attack opponents, RTAA and SPARK, are not applicable as a white-box attack on new transformer and non-transformer trackers containing MixFormerM, OSTrackSTS and RTS. For the RTS tracker~\citep{bhat_learning_2020}, though, TrackPGD is the only white-box attack which is applicable and significantly affect the tracker performance. 

The TransT-SEG~\citep{chen_high-performance_2023} and RTS~\citep{paul_robust_2022} trackers robustness after a white-box attack is also assessed on the UAV123 dataset in Table~\ref{tab:uav}. Once again, our method achieves scores that are closely competitive with SPARK~\citep{guo_spark_2020} and RTAA~\citep{jia_robust_2020} attacks on bounding box prediction, even though TrackPGD is developed to mislead trackers in predicting binary masks. Also, our method is the first white-box attack applicable on RTS tracker and presents a very effective performance in decreasing the RTS scores on the UAV123~\citep{mueller_benchmark_2016} dataset. To check further examples of bounding box prediction after attacks, we refer the reader to the Supplementary materials.

\begin{table}
    \centering  
    \caption{TransT-SEG~\citep{chen_high-performance_2023} and RTS~\citep{paul_robust_2022} trackers evaluation after white-box applicable approaches, SPARK~\citep{guo_spark_2020}, RTAA~\citep{jia_robust_2020} and TrackPGD (ours), on the GOT10k~\citep{huang_got-10k_2021} dataset.} \label{tab:got}
    \begin{tabular}{p{0.14\textwidth}p{0.2\textwidth}p{0.1\textwidth}p{0.1\textwidth}p{0.1\textwidth}} \\ 
    \toprule
     Tracker & Attacker & AO & SR$_{0.5}$ & SR$_{0.75}$ \\
    \midrule 
     \multirow{4}{*}{{TransTSEG}}  
     & No Attack  & 0.719 & 0.816 & 0.680  \\
      & RTAA & 0.127 & 0.115 & 0.060  \\
      & SPARK & 0.120 & 0.068 & 0.015  \\
      & TrackPGD (ours) & 0.139 & 0.092 & 0.025 \\ [0.15cm]
      \midrule 
      \multirow{2}{*}{{RTS}}  
     & No Attack  & 0.736 & 0.830 & 0.681  \\
      & TrackPGD (ours) & 0.292 & 0.291 & 0.162 \\
       \bottomrule
    \end{tabular} 
    \end{table}

\begin{table}
    \centering 
        \caption{Evaluation of our proposed TrackPGD vs. other white-box attacks applied on TransT-SEG~\citep{chen_high-performance_2023} and RTS~\citep{paul_robust_2022} trackers on the UAV123 dataset.}
    \label{tab:uav}
    \begin{tabular}{p{0.1\textwidth}p{0.12\textwidth}p{0.1\textwidth}p{0.1\textwidth}} \\ 
    \toprule
    Tracker & Attacker & AUC & Precision \\
    \midrule 
    \multirow{4}{*}{{TranstSEG}}  
     &  No attack & 0.670 & 0.859 \\
     &   SPARK & 0.050 & 0.074 \\
     &   RTAA &  0.058 &  0.077 \\
     &   TrackPGD & 0.135 & 0.205 \\
    \midrule
    \multirow{2}{*}{{RTS}}  
     & No attack & 0.666 & 0.873   \\
     & TrackPGD &  0.427 & 0.674   \\
       \bottomrule
    \end{tabular} 
\end{table}

\section{Conclusion}

Existing adversarial attacks for object trackers are ineffective as white-box attacks against transformer trackers. To evaluate the adversarial robustness of transformer trackers, new white-box attacks that utilize the tracker's own gradients are necessary. This paper introduces TrackPGD, a novel white-box attack that leverages object binary masks to assess the adversarial robustness of transformer trackers. We highlight the effectiveness of our proposed difference loss in impacting tracker performance compared to standard segmentation losses such as SegPGD. The efficacy of TrackPGD is validated through comprehensive experiments on various transformer and non-transformer networks on popular datasets. Despite the strong performance of our method, TrackPGD is limited to object trackers that generate a binary mask. Future work could extend these attacks to a broader range of trackers. Additionally, exploring defense strategies, such as adversarial purification with generative models (e.g., GANs or diffusion models) or adversarial training, could improve adversarial robustness in object tracking networks.

\subsubsection*{Acknowledgments}

This work is supported by the DEEL Project CRDPJ 537462-18 funded by the Natural Sciences and Engineering Research Council of Canada (NSERC) and the Consortium for Research and Innovation in Aerospace in Québec (CRIAQ), together with its industrial partners Thales Canada inc, Bell Textron Canada Limited, CAE inc and Bombardier inc. \footnote{\url{https://deel.quebec}}


\small

\bibliography{bibliography}


\appendix

\section{Supplementary materials}

\subsection{Algorithm}


 
The principle steps of TrackPGD algorithm is presented in Algorithm~\ref{alg:TPGD}. The sign of step size $\alpha$, line 9 in Algorithm~\ref{alg:TPGD}, is determined according the best performance of each tracker, following the decoupling of the norm and direction of gradients in adversarial attacks~\citep{rony_decoupling_2019}.

\begin{algorithm*}
\footnotesize
\caption{TrackPGD to attack transformer trackers with segmentation capability}\label{alg:TPGD}
\begin{algorithmic}[1]
\Require Tracker $\mathcal{F}(\cdot)$, current frame ${I}_{\tau}$, previous binary mask ${M}_{\tau-1}$, perturbation range $\epsilon$, step size $\alpha$, loss trade-offs $\lambda_1$ and $\lambda_2$, maximum iteration ${T}$, focusing parameter $\gamma$, variant of focal loss $\alpha_t$, probability map $p_t$   \smallskip  
\State ${I}_\text{adv}^0 \gets I_{\tau}$ \Comment{initialization} \smallskip
\State $G_{\tau} \gets M_{\tau-1}$ \Comment{use last predicted binary mask as ground truth} 
\For{$t = 1\ldots T$} 
\State ${M}^{t} \gets \mathcal{F}({I}_\text{adv}^{t-1})$ \Comment{predict binary mask} \smallskip
\State $L_{\Delta}  \gets  L_\text{SegPGD}(M^t, G_{\tau}) - L_\text{SegPGD}(M^t, 1-G_{\tau})  $ 
\Comment{compute difference of SegPGD losses} 
\smallskip
\State $L_\text{focal} \gets \alpha_t  (1 - p_t)^{\gamma}  L_{\Delta}$ 
\Comment{compute focal loss} 
\smallskip
\State $L_\text{dice} \gets 1 - 2\,\text{IoU}(M^t, G_{\tau})$ 
\Comment{compute dice loss} 
\smallskip
\State $L \gets \lambda_1  L_\text{focal} + \lambda_2 L_\text{dice}$ \Comment{compute TrackPGD loss} \smallskip
\State ${I}_\text{adv}^t \gets I_\text{adv}^{t-1} +  \alpha\,\mathrm{sign}(\nabla_{I_\text{adv}^{t-1}} L)$ 
\Comment{update adversarial example} 
\smallskip
\State ${I}_\text{adv}^t \gets \phi^{\epsilon} \left({I}_\text{adv}^t\right) $ \Comment{clip to the $\epsilon$-ball } 
\EndFor 
\end{algorithmic}
\end{algorithm*}

\subsection{Ablation study}


\label{sec:intuition}
\begin{figure}
    \centering \tiny
     \caption{The before (green) and after attack masks (red) generated by MixFormerM~\citep{cui_MixFormer_2022}, while the vanilla SegPGDs and difference loss are used instead of binary cross entropy in the focal loss for the TrackPGD attack. } 
     \label{fig:vis}
	\begin{tabular}{p{0.24\textwidth}p{0.24\textwidth}p{0.24\textwidth}} 
		$\quad L_{\text{SegPGD}} (M^t, G_\tau)$ & $ -L_{\text{SegPGD}} (M^t, 1 - G_\tau)$ & $\qquad \quad L_\Delta$ \\ 
        \centerline{\includegraphics[width=0.24\textwidth]{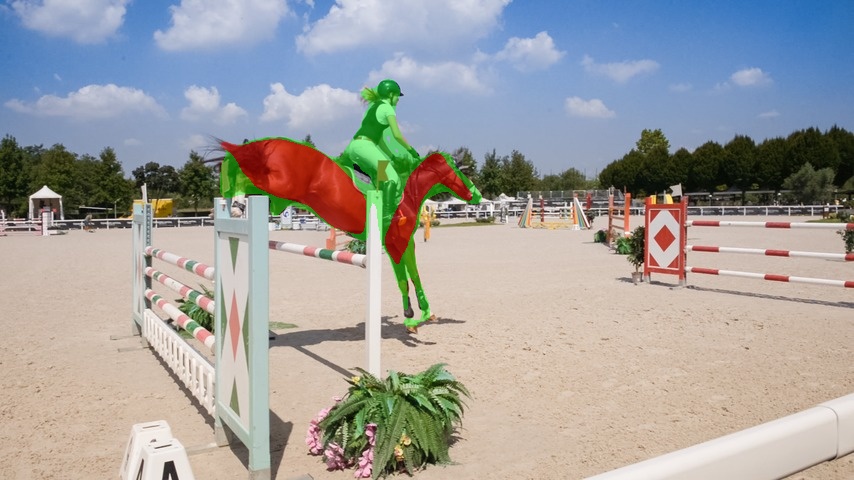}} &
         \centerline{\includegraphics[width=0.24\textwidth]{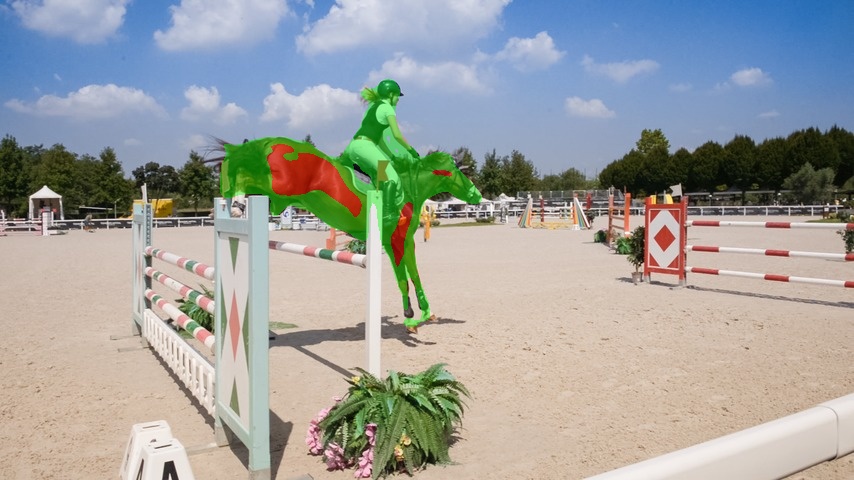}}  & 
         \centerline{\includegraphics[width=0.24\textwidth]{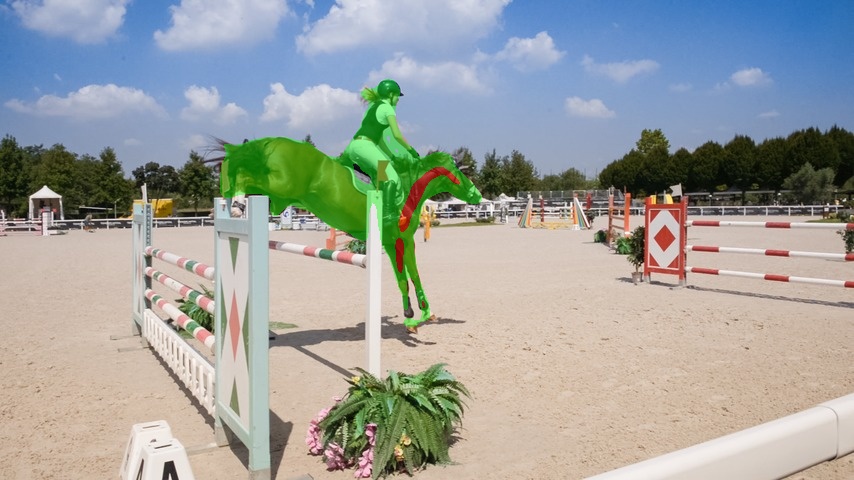}} \\ 
            \centerline{\includegraphics[width=0.24\textwidth]{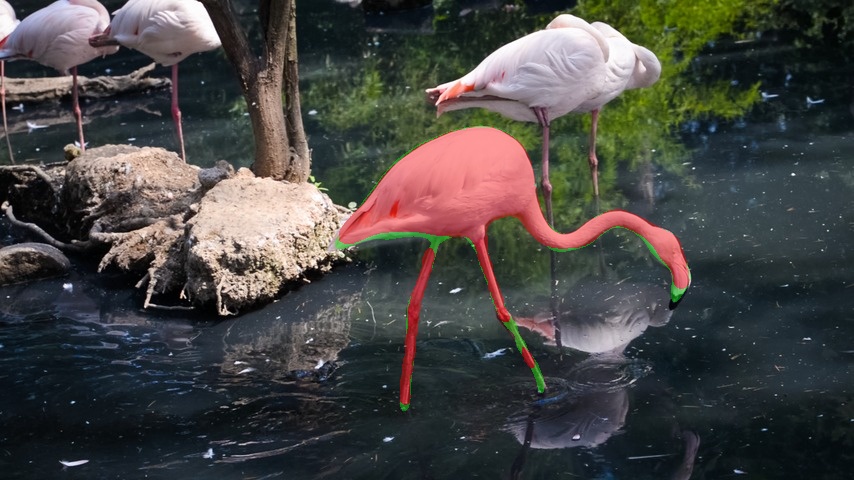}} & 
            \centerline{\includegraphics[width=0.24\textwidth]{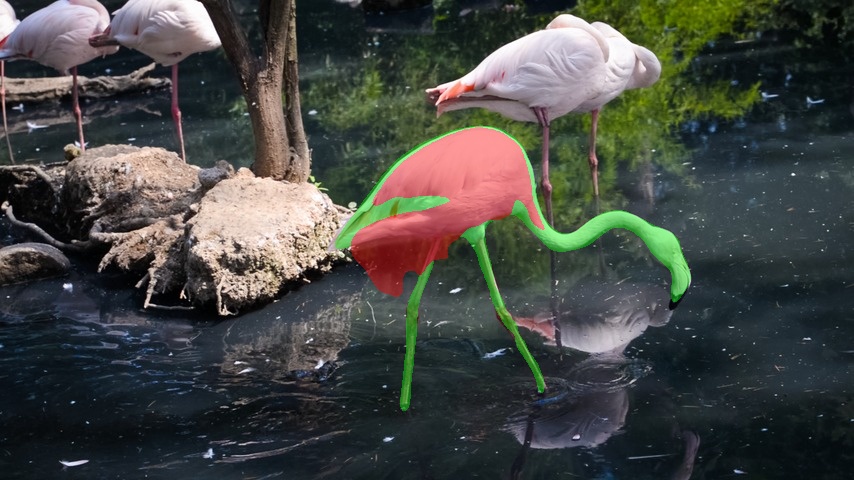}} & 
            \centerline{\includegraphics[width=0.24\textwidth]{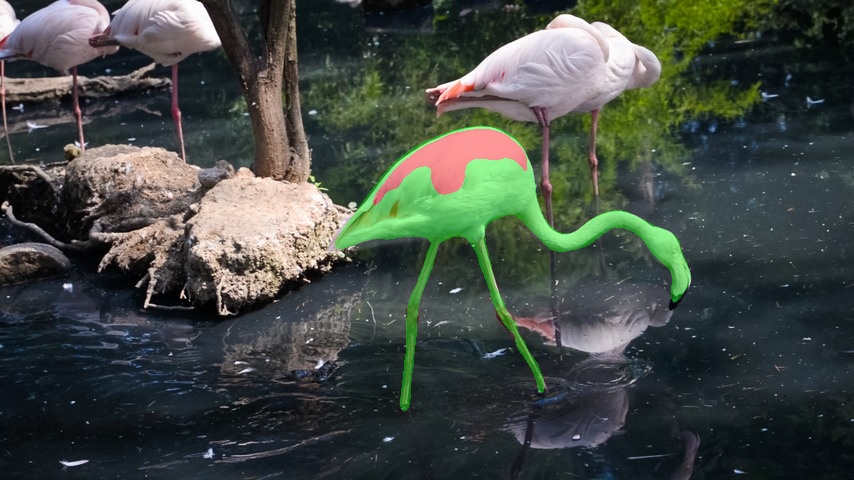}} 
	\end{tabular} 
\end{figure}

\paragraph{Role of $L_{\Delta}$ in TrackPGD} Inspired by the SegPGD attack~\citep{gu_segpgd_2022}, TrackPGD relies on the predicted binary mask to generate the adversarial frames. The main challenge in following SegPGD for binary mask impairment to attack object trackers is that the SegPGD works well as a multi-class segmentation attack; while here, we have only two classes to switch the pixel labels. The object class and the background class pixels should switch their labels after TrackPGD. However, the object pixels are limited in comparison to the background pixels in almost all of the video frames. For this specific reason, it is necessary to define our method based on both object and background masks to involve all of the frame pixels in generating the adversarial perturbation. Plus, given the availability of two classes in the tracking scenario, we customized the segmentation loss using the tracking loss of the TransT~\citep{chen_transformer_2021} to effectively attack binary masks. We also demonstrate the role of ${L}_{\Delta}$ in TrackPGD versus using the vanilla SegPGD losses for the object and background masks, in Figure~\ref{fig:vis}. The after attack masks show the essential role of our proposed combination loss $L_{\Delta}$ in crashing the object mask inferred by MixFormerM~\citep{cui_MixFormer_2022}. Although, the vanilla SegPGD losses also generate inaccurate binary masks, the mask impairment caused by $L_{\Delta}$ is significantly greater.

\begin{table}
    \centering \footnotesize
    \caption{The impact of cross-entropy substitutions in focal loss on $J(M)\%$. The vanilla SegPGD losses versus the difference loss in the TrackPGD against MixFormerM~\citep{cui_MixFormer_2022} on DAVIS2016~\citep{mueller_benchmark_2016}.}
    \label{tab:abL}
    \begin{tabular}{p{0.15\textwidth}p{0.22\textwidth}p{0.25\textwidth}p{0.13\textwidth}} \\
     \toprule
     Original & $L_{\text{SegPGD}}(M^t, G_\tau)$ & $- L_{\text{SegPGD}}(M^t, 1- G_\tau)$ & $L_{\Delta}$  \\
    \midrule
     85.82 & 52.86 & 37.28 & \textbf{30.30} \\
       \bottomrule
    \end{tabular}
\end{table}

In addition, we conducted an ablation study to demonstrate the difference loss role in the focal loss. We predicted the binary masks by the MixFormerM tracker attacked by TrackPGD where the binary cross entropy is substituted with the vanilla SegPGD loss for the object mask, vanilla SegPGD for the background mask, and the difference loss $L_{\Delta}$. The DAVIS2016~\citep{perazzi_benchmark_2016} dataset was used to compute the percentage of the mean Jaccard index $J(M)$ of the binary masks indicating the average overlap between predicted and annotated binary masks. TrackPGD attack is stronger with the proposed difference loss, see Table~\ref{tab:abL}, in comparison with the vanilla SegPGD losses for the object mask $ G_\tau$, and the background mask $1 -  G_\tau$. The difference loss presented the most effective attack based on the generated binary masks by MixFormerM~\citep{cui_MixFormer_2022}, which has the smaller $J(M)\%$ on average.

\paragraph{Coefficients of the TrackPGD Loss} In this study, we show the effect of each loss term, dice and focal losses, in the TrackPGD attack. We conducted experiments on DAVIS2016~\citep{perazzi_benchmark_2016} using two evaluation metrics: the mean Jaccard index $J(M)$ and contour accuracy $F(M)$ of the predicted masks. The results were presented in Table~\ref{tab:tablation} where the average $J\&F(M)$ percentage was used to demonstrate the impact of each losses term, focal vs.\ dice loss. Table~\ref{tab:tablation} shows that the focal loss is more effective than the dice loss, resulting in smaller $J\&F(M)$ values. The $J\&F(M)$ value is used to measure the effectiveness of the attack, where a smaller value represents a more significant influence. We selected powers of 10 as coefficients of focal loss and powers of 2 as coefficients of dice loss. As in the literature, the loss related to IoU is normally 1 or 2~\citep{chen_high-performance_2023,cui_MixFormer_2022}, while the loss terms related to binary cross entropy (or focal loss) are selected as 1000~\citep{yan_alpha-refine_2021} or 1~\citep{chen_high-performance_2023}. An interesting observation in Table~\ref{tab:tablation} is that the MixFormerM is significantly impacted by TrackPGD, more than the TransT-SEG tracker using the same set of experiments. In both tables, the MixFormerM tracker achieves the smallest values of $J\&F(M)$ meaning more significant attack results.

\begin{table}[b]
    \centering \footnotesize
    \caption{$J\&F(M)\%$ values with varying focal and dice losses. Present the role of focal loss (changing $\lambda_1$ when $\lambda_2=0$) and dice loss (changing $\lambda_2$ when $\lambda_1=0$) in the TrackPGD attack. $J\&F(M)\%$ values are computed on DAVIS2016~\citep{mueller_benchmark_2016} dataset.}
    \label{tab:tablation}
    \begin{tabular}{p{0.2\textwidth}p{0.07\textwidth}p{0.07\textwidth}p{0.07\textwidth}p{0.07\textwidth}p{0.07\textwidth}p{0.07\textwidth}} 
    \midrule
       $\lambda_1$ ($\lambda_2=0$) & $10000$ & $1000$  & $100$ & 
        $10$ & $1$ & $0.1$ \\
    \midrule 
         { MixFormerM} & 27.97 & 27.87 & 27.76 & 27.90 & \textbf{27.73} & 27.93 \\
         { TransT-SEG} & 28.98 & \textbf{28.52} & 28.74 & 
         28.71 & 28.65 &28.67 \\
    \midrule
       $\lambda_2$ ($\lambda_1=0$) & $4$ & $2$  & $1$ & $0.5$ & $0.25$ & $0.125$ \\
        \midrule 
         { MixFormerM} & 30.00 & 30.27 & \textbf{29.92} & 30.04 & 30.14 & 29.92\\
         {TransT-SEG} & 33.62 &  33.62 & 33.52 & 34.05 & \textbf{33.14} & 33.55 \\
       \bottomrule
    \end{tabular}
    
\end{table}

\subsection{Fine tuning the parameters}

We used the DAVIS2016 dataset as the validation set to adjust the $ \lambda_1$ and $\lambda_2 $ parameters. The DAVIS protocol calculates the Jaccard Index \( \mathcal{J} \) and contour accuracy \( \mathcal{F} \) via three error statistic measures: mean, recall, and decay. The Jaccard index measures the Intersection over Union (IoU) metric, while the contour accuracy is computed by calculating the F-measure of the predicted mask and the ground truth.  To obtain the F-measure, the predicted mask is considered as a set of closed contours, and precision \( P_c \) and recall \( R_c \) are calculated between contour points by comparing these contours with the ground-truth contours. Then, the F-measure \( \frac{2P_c R_c}{P_c + R_c} \) is computed to represent the accuracy of the predicted mask. To fine-tune the TrackPGD parameters, we computed the average value of \( \mathcal{J}(R) \) and \( \mathcal{F} (R) \), represented as \( \mathcal{J}\&\mathcal{F} (R) \). This approach considers both IoU and contour accuracy over time across video frames. Since this measurement is close to the robustness definition in VOT challenge~\citep{kristan_tenth_2023}, we chose this metric to identify the best hyperparameters to determine the optimal coefficient of TrackPGD loss.

\begin{figure*}
     \centering
     \includegraphics[width=0.44\textwidth]{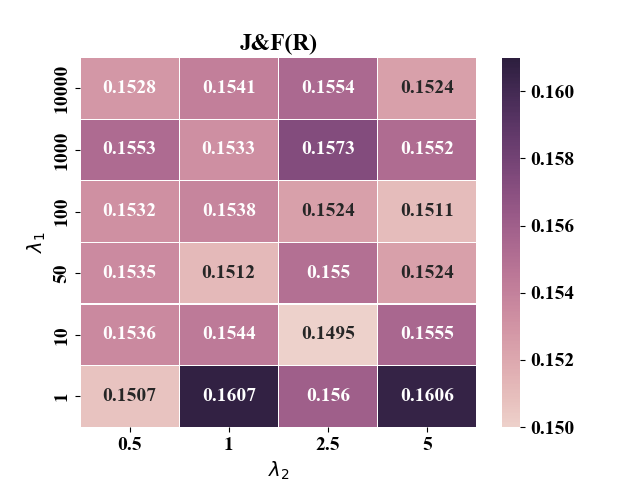} 
     \includegraphics[width=0.44\textwidth]{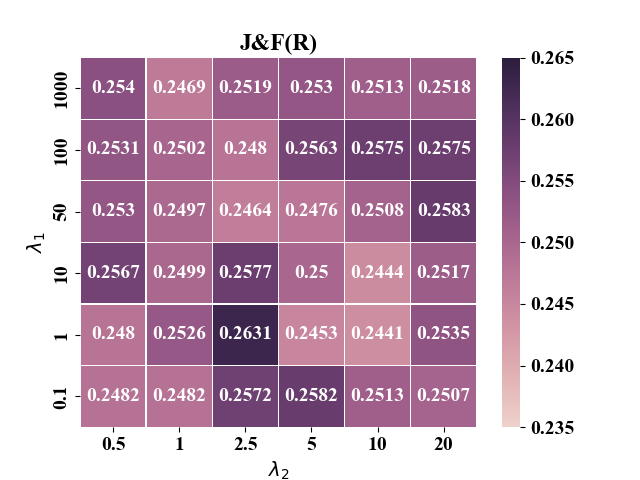}
     \includegraphics[width=0.44\textwidth]{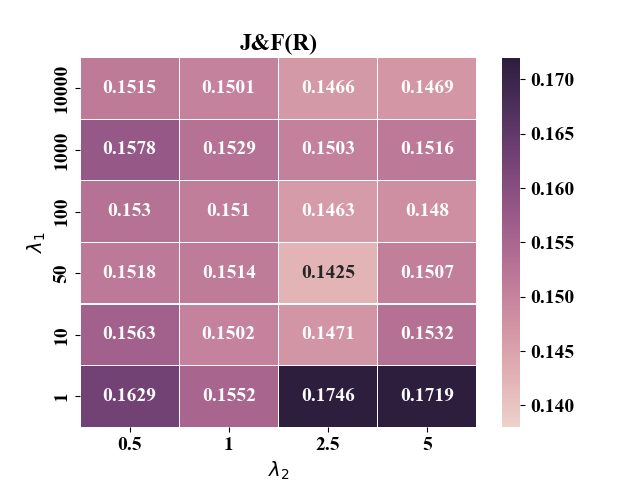}
     \includegraphics[width=0.44\textwidth]{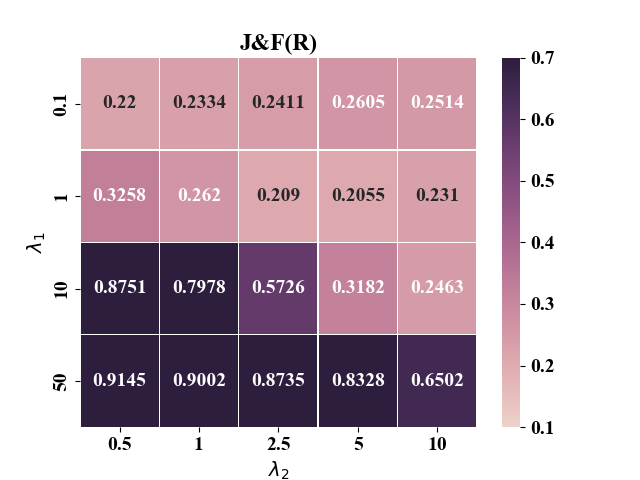}
     \caption{The TrackPGD parameter heatmaps from left to right, first row: MixFormerM\citep{cui_MixFormer_2022, yan_alpha-refine_2021}, OSTrackSTS~\citep{ye_joint_2022, yan_alpha-refine_2021} and second row: TransT-SEG~\citep{chen_high-performance_2023}, RTS~\citep{paul_robust_2022} trackers.}
     \label{fig:heatmaps}
\end{figure*}

The following heatmaps show the average values of the Jaccard index \( \mathcal{J} \) and contour accuracy \( \mathcal{F} \) for the tracker's recall metric. The better performance of the tracker is represented by the greater \( \mathcal{J}\&\mathcal{F} (R) \). As our reported numbers are based on attack results, a smaller number indicates a more effective attack. Therefore, the best parameters for MixFormerM are \( ( \lambda_1, \lambda_2)  = (10, 2.5) \), while the TrackPGD is strongest against OSTrackSTS with \( (\lambda_1, \lambda_2) = (1, 10) \), against TransT-SEG with \( (\lambda_1, \lambda_2) = (50, 2.5) \) and against RTS with \((\lambda_1, \lambda_2) = (1, 5) \). It is also worth mentioning that optimal performance is achieved when both terms are active at intermediate values, as shown in Figure~\ref{fig:heatmaps}. This suggests that neither term alone yields the best results.

\subsection{Binary mask assessment}

\begin{figure}
     \centering
     \centerline{\includegraphics[width=\textwidth]{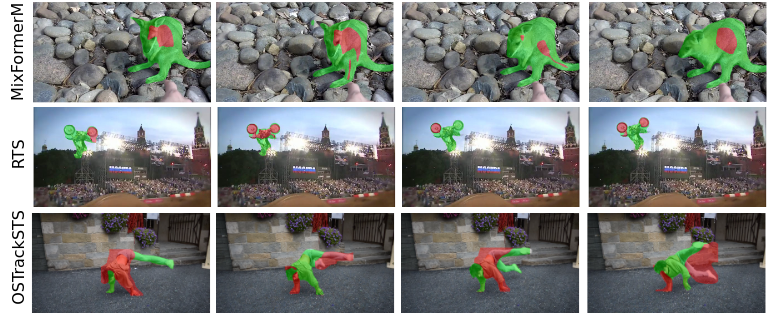}}
     \caption{The results of MixFormerM~\citep{cui_MixFormer_2022}, RTS~\citep{paul_robust_2022}, and OSTrackSTS~\citep{ye_joint_2022} attacked by TrackPGD. The green mask is the original output, while the red mask is the after attack mask.}
     \label{fig:msk}
\end{figure}

\paragraph{Qualitative evaluation.} Figure~\ref{fig:msk} represents the performance of TrackPGD in generating inaccurate binary masks for MixFormerM~\citep{cui_MixFormer_2022} and OSTrackSTS~\citep{ye_joint_2022} and RTS~\citep{paul_robust_2022} trackers on the successive frames of several video sequences. The green color represents the original trackers' output, while the red color represents the tracker output after the TrackPGD attack. As it is presented in Table~\ref{tab:votMT}, the effect of the TrackPGD attack on the MixFormerM performance is more significant.

\subsection{Bounding box assessment}

\begin{figure}
     \centering
     \centerline{\includegraphics[width=0.98\textwidth]{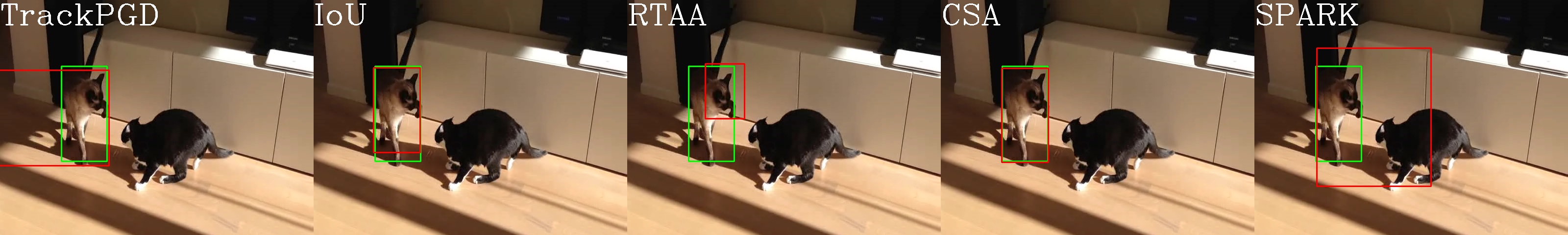}}
     \centerline{\includegraphics[width=0.98\textwidth]{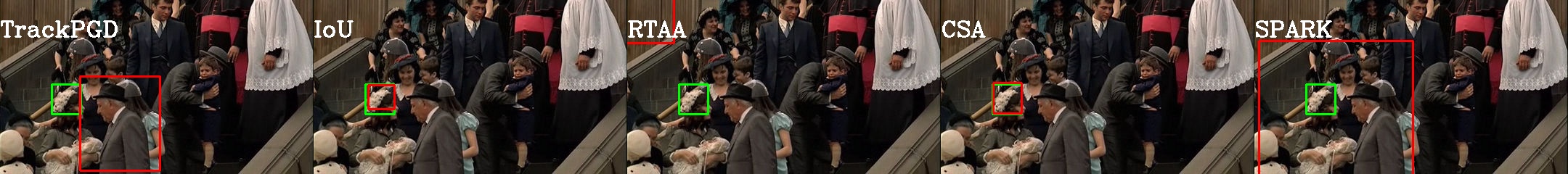}}
     \centerline{\includegraphics[width=0.98\textwidth]{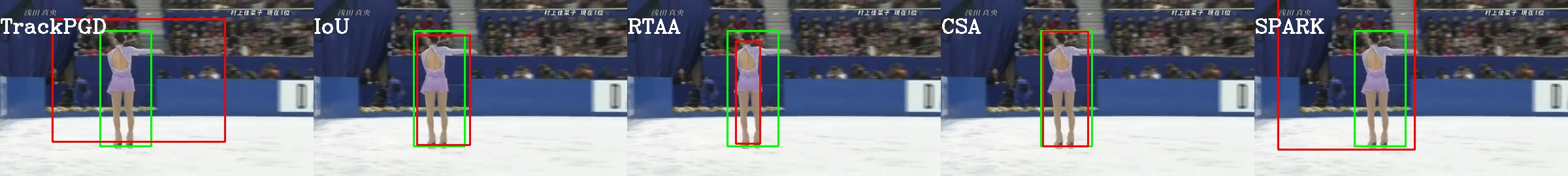}}
     \centerline{\includegraphics[width=0.98\textwidth]{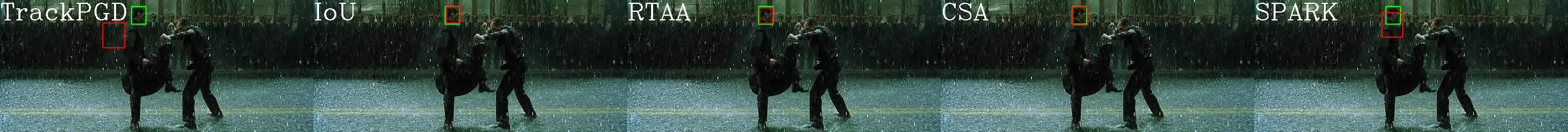}}
     \caption{The TransT-SEG~\citep{chen_high-performance_2023} tracker after SPARK~\citep{guo_spark_2020}, RTAA~\citep{jia_robust_2020}, CSA~\citep{yan_cooling-shrinking_2020}, IoU~\citep{jia_iou_2021} and TrackPGD. The green/red colors correspond to before and after the attack.}
     \label{fig:box}
\end{figure}

\paragraph{Quantitative evaluation.} Table~\ref{fig:box} presents the attack results on the TransT-SEG~\citep{chen_high-performance_2023} and RTS~\citep{paul_robust_2022} trackers using the GOT-10k dataset~\citep{huang_got-10k_2021}. Although our method focuses on manipulating object binary masks, the reported numbers for bounding box evaluation are very close to those of competitive attacks aimed at misleading trackers into providing inaccurate bounding boxes. We surpassed the RTAA attack on the GOT-10k dataset in both success rate scores. However, RTAA performs better in terms of average overlap (AO). TrackPGD scores are also very close to SPARK scores, the most competitive attack using object bounding boxes on GOT-10k~\citep{huang_got-10k_2021}, in misleading trackers to predict accurate object bounding boxes. Despite our method's aim to misclassify object and background pixels, it proves to be a highly competitive approach compared to bounding box-centered attacks. Furthermore, our TrackPGD has an extra advantage over other white-box attacks which is that other attack opponents, RTAA and SPARK, are not applicable as a white-box attack on new transformer trackers containing MixFormerM, OSTrackSTS and RTS. For the RTS tracker~\citep{bhat_learning_2020}, though, TrackPGD is the only white-box attack which is applicable and significantly affect the tracker performance. 

The TransT-SEG~\citep{chen_high-performance_2023} and RTS~\citep{paul_robust_2022} trackers robustness after a white-box attack is also assessed on the UAV123 dataset in Table~\ref{tab:uav}. Once again, our method achieves scores that are closely competitive with SPARK~\citep{guo_spark_2020} and RTAA~\citep{jia_robust_2020} attacks on bounding box prediction, even though TrackPGD is developed to mislead trackers in predicting binary masks. Also, our method is the first white-box attack applicable on RTS tracker and presents a very effective performance in decreasing the RTS scores on the UAV123~\citep{mueller_benchmark_2016} dataset.

\begin{table}
    \centering  \footnotesize
    \caption{TransT-SEG~\citep{chen_high-performance_2023} and RTS~\citep{paul_robust_2022} trackers evaluation after white-box applicable approaches, SPARK~\citep{guo_spark_2020}, RTAA~\citep{jia_robust_2020} and TrackPGD (ours), on the GOT10k~\citep{huang_got-10k_2021} dataset.} \label{tab:got}
    \begin{tabular}{p{0.14\textwidth}p{0.2\textwidth}p{0.1\textwidth}p{0.1\textwidth}p{0.1\textwidth}} \\ 
    \toprule
     Tracker & Attacker & AO & SR$_{0.5}$ & SR$_{0.75}$ \\
    \midrule 
     \multirow{4}{*}{{TransTSEG}}  
     & No Attack  & 0.719 & 0.816 & 0.680  \\
      & RTAA & 0.127 & 0.115 & 0.060  \\
      & SPARK & 0.120 & 0.068 & 0.015  \\
      & TrackPGD (ours) & 0.139 & 0.092 & 0.025 \\ [0.15cm]
      \midrule 
      \multirow{2}{*}{{RTS}}  
     & No Attack  & 0.736 & 0.830 & 0.681  \\
      & TrackPGD (ours) & 0.292 & 0.291 & 0.162 \\
       \bottomrule
    \end{tabular} 
    \end{table}

\begin{table}
    \centering \footnotesize
        \caption{Evaluation of our proposed TrackPGD vs. other white-box attacks applied on TransT-SEG~\citep{chen_high-performance_2023} and RTS~\citep{paul_robust_2022} trackers on the UAV123 dataset.}
    \label{tab:uav}
    \begin{tabular}{p{0.1\textwidth}p{0.12\textwidth}p{0.1\textwidth}p{0.1\textwidth}} \\ 
    \toprule
    Tracker & Attacker & AUC & Precision \\
    \midrule 
    \multirow{4}{*}{{TranstSEG}}  
     &  No attack & 0.670 & 0.859 \\
     &   SPARK & 0.050 & 0.074 \\
     &   RTAA &  0.058 &  0.077 \\
     &   TrackPGD & 0.135 & 0.205 \\
    \midrule
    \multirow{2}{*}{{RTS}}  
     & No attack & 0.666 & 0.873   \\
     & TrackPGD &  0.427 & 0.674   \\
       \bottomrule
    \end{tabular} 
\end{table}

 \paragraph{Qualitative evaluation.} We also illustrate the bounding box evaluation of TransT-SEG tracker~\citep{chen_high-performance_2023} in several frames from the VOT2018~\citep{kristan_sixth_2018} sequence in Figure~\ref{fig:box}. The greed/red colors indicates the before and after attack predictions. Our proposed method, TrackPGD, results in misleading the tracker while its perturbation is generated only based on the binary mask. Other attacks such as SPARK and RTAA, which fully built upon bounding box(es) also significantly degrade the target bounding boxes. As the quantitative results in Table~\ref{tab:got} indicated, the IoU~\citep{jia_iou_2021} and CSA~\citep{yan_cooling-shrinking_2020} attack, black-box attacks, are behind the white-box methods' performance.  
 
\end{document}